\documentclass[journal]{IEEEtran}

\usepackage{graphicx}
\usepackage{cite}
\usepackage[T1]{fontenc}
\usepackage{amsmath,amssymb}
\usepackage{booktabs}
\usepackage[table]{xcolor}
\usepackage{multirow}
\usepackage{tabularx}
\usepackage{siunitx}
\usepackage{etoolbox}
\usepackage[caption=false,font=footnotesize]{subfig}
\usepackage[hidelinks]{hyperref}
\usepackage{capt-of}
\usepackage{placeins}
\usepackage{dblfloatfix}

\graphicspath{{figures/}}
\newcommand{\method}{\textsc{CalibSwitch}}
\newcommand{\generator}{\textsc{CAEG}}
\newcommand{\policy}{\textsc{SRCP}}
\newcommand{\SE}{\mathrm{SE}(3)}
\newcommand{\SO}{\mathrm{SO}(3)}

\definecolor{GainGreen}{RGB}{8,100,67}
\definecolor{OursTint}{RGB}{237,246,239}
\newcommand{\gdown}[1]{{\color{GainGreen}\scriptsize$\downarrow\,#1\%$}}
\newcommand{\gup}[1]{{\color{GainGreen}\scriptsize$\uparrow\,#1$\,pts}}
\newcommand{\gaincell}[2]{\mbox{\textbf{#1}\ifstrempty{#2}{}{\,{\color{GainGreen}\scriptsize(#2)}}}}
\newcolumntype{G}[1]{>{\hsize=#1\hsize\linewidth=\hsize\centering\arraybackslash}X}

\newcommand{\highErrorNDSGain}{5.37}

\hypersetup{pdftitle={CalibSwitch: Observation-Driven Camera–LiDAR Extrinsic Recovery and Selective Maintenance for Multimodal Perception},pdfauthor={Haifa Zhang, Yijing Wang, Zhiqiang Zuo},pdfsubject={Research preprint}}
\begin{document}

\title{CalibSwitch: Observation-Driven Camera--LiDAR Extrinsic Recovery and Selective Maintenance for Multimodal Perception}

\author{Haifa~Zhang, Yijing~Wang, and Zhiqiang~Zuo
\thanks{The authors are with the School of Electrical and Information Engineering, Tianjin University, Tianjin, China.}}

\maketitle
\thispagestyle{plain}
\pagestyle{plain}

\begin{abstract}
Camera--LiDAR perception on mobile robots and intelligent vehicles depends on reliable sensor geometry throughout deployment. Installation variations create uncertainty in the initial extrinsics, while changing observation quality makes frame-wise estimates fluctuate during operation. This paper presents CalibSwitch, an observation-driven framework that separates absolute extrinsic recovery from selective state maintenance. A cross-modal absolute extrinsic generator (CAEG) combines projection-free bidirectional feature interaction with geometry-regularized learning to recover complete SE(3) candidates from image--point observations and coarse view context, without a supplied numerical initial transform. A selective recurrent calibration policy (SRCP) combines observation quality and temporal history to determine when a candidate should replace the working geometry consumed by a frozen multimodal detector. On nuScenes, CAEG achieves mean translation and rotation errors of 0.248 m and 5.884$^\circ$. Under sustained geometric transitions, SRCP reaches a mean composite error of 1.46 with 9.70\% candidate commitments, outperforming quality-based and temporal-consistency gating at comparable commitment rates. Relative to unconditional per-frame updating, it improves frozen-detector mAP and NDS by 1.43 and 0.92 points. Paired geometry--observation uncertainty tests, two real-vehicle platforms, and 13.5-FPS single-camera Jetson execution demonstrate the system value of selective geometric-state maintenance for continuous perception.
\end{abstract}


\section{Introduction}
\label{sec:introduction}

\IEEEPARstart{C}{amera--LiDAR} fusion combines image semantics with metric spatial information for perception on mobile robots, intelligent vehicles, and industrial sensing platforms~\cite{liu2023bevfusion,bai2022transfusion,li2022deepfusion}. Its operation relies on the geometric relationship between physically separated sensors. Practical systems undergo independent installation, assembly adjustment, sensor replacement, and platform transfer; the actual working configuration can consequently differ from the design or previously calibrated configuration~\cite{liu2023highprecision}. Structural and operational changes add further uncertainty during deployment~\cite{li2024twostep,wu2026consistency}. These engineering conditions motivate a system that can establish usable geometry from current observations and maintain that geometry throughout continuous operation.

Geometric adaptation and reliable operation act on different aspects of this problem. A persistent configuration change calls for a new working extrinsic. Meanwhile, illumination, occlusion, scene structure, and LiDAR sparsity change the reliability of instantaneous estimates~\cite{yu2023robustness,beemelmanns2024multicorrupt}. A large difference between a new estimate and the maintained state may therefore reflect either a lasting geometric change or a transient observation disturbance. The system must use the available sensing evidence and its history to decide when to adapt. This motivates treating extrinsics as maintained perception states whose acceptance is governed by observation reliability.

Existing calibration recovers sensor geometry through engineered targets, natural-feature alignment, motion constraints, or learned cross-modal representations~\cite{liu2020onsite,an2019omnidirectional,li2024twostep}. Residual learning efficiently corrects a supplied transform within a learned neighborhood~\cite{schneider2017regnet,iyer2018calibnet,lv2021lccnet,xiao2024calibformer}, while target-free alignment and consistency-aware estimation recover geometry through scene or multiview evidence~\cite{daddeo2026calibbev,han2026unicalib,wu2026consistency}. Online estimation further exploits trajectories and finite histories~\cite{shi2020calibrcnn,lee2025finite}; event-triggered fusion uses information-dependent update decisions to support continuous operation~\cite{cao2026event,liu2023rlmpc}. Together, these developments motivate two coupled operations for deployment: generating a complete candidate from current multimodal evidence and deciding whether that candidate should become the geometry used by the perception system.

The key insight of \method{} is that generating a geometric estimate and committing it to a working perception state are distinct operations. The candidate $\widehat T_t$ represents geometry supported by current observations, whereas the maintained extrinsic $T_t$ determines the geometry consumed by subsequent perception. Each commitment therefore affects both the current frame and the reliability of the working state over time. Jointly considering instantaneous observation quality and historical geometric consistency enables adaptation to persistent configuration changes while suppressing transient estimation disturbances. This persistent effect motivates recurrent decisions and sequence-level learning.

We implement this idea through an observe--generate--select--detect chain. The cross-modal absolute extrinsic generator (\generator) uses camera images, LiDAR points, and coarse view context to recover complete camera-to-LiDAR transforms without a numerical initial matrix. Projection-free bidirectional feature interaction establishes joint image--point evidence, and geometry-regularized supervision connects pose parameters to rigid-body action and spatial consistency. The selective recurrent calibration policy (\policy) selects \textsc{keep} or \textsc{update}, and a frozen BEVFusion detector consumes the resulting working state. Evaluation follows the same progression: initial recovery, selective maintenance, downstream perception, and natural deployment on two real platforms. The contributions are as follows.

\begin{itemize}
    \item We develop an observation-driven absolute extrinsic generator that combines bidirectional image--point interaction and geometry-regularized supervision to recover complete camera--LiDAR transformations, conditioned on coarse view context and without a supplied numerical initial matrix.
    \item We formulate online calibration maintenance as a recurrent state-commitment problem. The proposed policy combines current observation quality and historical consistency to adapt to persistent geometric changes while retaining reliable geometry during transient observation disturbances.
    \item We connect recovery and selective maintenance to a frozen multimodal detector and establish their system value through pose recovery, joint uncertainty tests and matched-commitment comparisons, mechanism analysis, two real-vehicle platforms, and embedded execution.
\end{itemize}

\section{Related Work}
\label{sec:related_work}

\subsection{Extrinsic Recovery under Deployment Uncertainty}

Extrinsic calibration establishes the geometric relationships required to deploy multisensor systems. In vision-guided robotics, joint calibration compensates for robot assembly, vision--robot alignment, and end-effector installation errors within a unified geometric model~\cite{liu2023highprecision}. On-site LiDAR--IMU calibration, omnidirectional color ranging, and continuous-time LiDAR/camera/IMU calibration extend this geometric integration to different sensing configurations~\cite{liu2020onsite,an2019omnidirectional,li2024twostep}. A recent survey organizes camera--LiDAR calibration around explicit and implicit cross-modal correspondence~\cite{an2024survey}, identifying how heterogeneous observations constrain the installed sensor relationship.

Target-free recovery obtains these constraints from the scene and sensor motion. Automatic single-shot calibration and cross-modal mask matching establish geometric associations from natural observations~\cite{koide2023toolbox,huang2025mask}. MDPCalib combines visual/LiDAR odometry with learned point--pixel correspondences in a joint optimization~\cite{petek2024mdpcalib}. TLC-Calib jointly optimizes extrinsics and a neural Gaussian scene representation~\cite{jung2026tlccalib}, while CalibBEV and UniCalib organize cross-modal alignment in BEV and unified depth representations~\cite{daddeo2026calibbev,han2026unicalib}. Consistency-aware robotic extrinsic estimation further couples recovered geometry to multiview reconstruction quality~\cite{wu2026consistency}. These approaches develop complementary sources of observation support for deployment-time recovery.

Learning-based calibration develops efficient mappings from cross-modal evidence to pose estimates. RegNet, CalibNet, LCCNet, and CalibFormer learn corrections to a supplied transform through feature regression, geometric supervision, cost volumes, or attention~\cite{schneider2017regnet,iyer2018calibnet,lv2021lccnet,xiao2024calibformer}. FusionNet uses hierarchical point--pixel fusion for coarse-to-fine recovery~\cite{wang2022fusionnet}; DXQ-Net models calibration-flow uncertainty and incorporates differentiable pose estimation to account for correspondence quality~\cite{jing2022dxq}. This progression makes the construction and reliability of cross-modal evidence central to geometric recovery. \generator{} uses projection-free bidirectional image--point interaction and geometry-regularized supervision to map current observations and coarse view context directly to a complete $\SE$ candidate, establishing the working geometry without a supplied numerical initial matrix.

\subsection{Online Calibration and State Maintenance}

Online calibration integrates repeated observations with an evolving estimate of system geometry. Motion-based calibration couples sensor trajectories with extrinsic and timing estimation~\cite{taylor2016motion}, and CalibRCNN combines recurrent features with geometric constraints~\cite{shi2020calibrcnn}. CalibDepth provides a complementary refinement mechanism through unified depth representations and iterative online calibration~\cite{zhu2023calibdepth}. For intelligent vehicles, concurrent learning compensates camera extrinsics while estimating structure and motion with an uncalibrated two-camera system~\cite{zhang2023uncalibrated}. Finite-memory simultaneous localization and calibration uses bounded measurement histories~\cite{lee2025finite}; inertial-aided recurrent EKF calibration applies repeated measurement updates to magnetometer calibration under magnetic interference~\cite{wang2025magnetometer}. Together, these studies establish historical evidence and recursive estimation as practical tools for maintaining calibration during operation.

The reliability of online recovery also depends on the support available in the observations. Motion excitation assessment guides data collection and initialization in target-free LiDAR--IMU--camera calibration~\cite{wang2025excitation}. Environment-driven calibration adapts feature extraction to scene feature density and combines structural and textural matching across views~\cite{huang2025environment}. At the estimate level, Cocheteux \emph{et al.} combine Monte Carlo dropout and conformal prediction to attach uncertainty intervals to online extrinsic estimates~\cite{cocheteux2025uncertainty}. These mechanisms connect candidate geometry to motion informativeness, cross-modal support, and estimation confidence. They provide a natural basis for deciding how a candidate should influence the geometry retained by a continuously operating perception system.

Selective state commitment extends this reliability assessment to the timing and persistent effect of an update. Event-triggered asynchronous fusion localization adapts when sensing information enters the state estimate~\cite{cao2026event}, while reinforcement-learning-based event selection jointly considers update timing and control performance~\cite{liu2023rlmpc}. Multisensor localization in dynamic and degraded environments likewise adapts its use of observation evidence~\cite{li2025simsl}. Building on temporal estimation and reliability assessment, \method{} distinguishes the current candidate $\widehat T_t$ from the maintained extrinsic $T_t$ consumed by subsequent perception. \policy{} combines present observation quality with recurrent history to select \textsc{keep} or \textsc{update}. This decision links the acceptance of current evidence to the reliability of the working geometric state over the following frames.

\subsection{Calibration-Aware Multimodal Perception}

Multimodal perception provides a task-level measure of the geometry supplied to sensor fusion. TransFusion associates image and point features through object queries, while BEVFusion combines modalities in a unified BEV representation~\cite{bai2022transfusion,liu2023bevfusion}. GAFusion adds LiDAR depth and occupancy guidance with adaptive global feature interaction~\cite{li2024gafusion}; cross-scene place recognition also exploits latent camera--LiDAR correlation~\cite{pan2025latent}. Robustness benchmarks relate spatial misalignment and observation degradation to detection quality~\cite{yu2023robustness,beemelmanns2024multicorrupt}, connecting geometric reliability to useful multimodal information. \method{} places recovery and selective commitment at this geometric interface. Joint evaluation of extrinsic error and frozen-detector performance traces how candidate recovery and maintained-state reliability improve continuous perception.

\section{Proposed CalibSwitch Framework}
\label{sec:method}

\subsection{Problem Formulation and Overview}

Let $\mathcal{I}_{t}=\{I_t^c\}_{c=1}^{C}$ and $P_t=\{(x_i,y_i,z_i,\rho_i)\}_{i=1}^{N}$ denote synchronized images and a LiDAR scan. Three variables define the deployment problem: the actual geometry $T_t^{c*}\in\SE$, the observation-dependent candidate $\widehat T_t^c$, and the maintained extrinsic $T_t^c$, a persistent geometric state explicitly committed for downstream perception. Installation uncertainty motivates complete candidate recovery; operational uncertainty motivates selective commitment. Reference transforms supervise recovery and policy learning; online operation uses observations, coarse view context, and maintained history.

\begin{figure*}[t]
    \centering
    \includegraphics[width=0.96\textwidth]{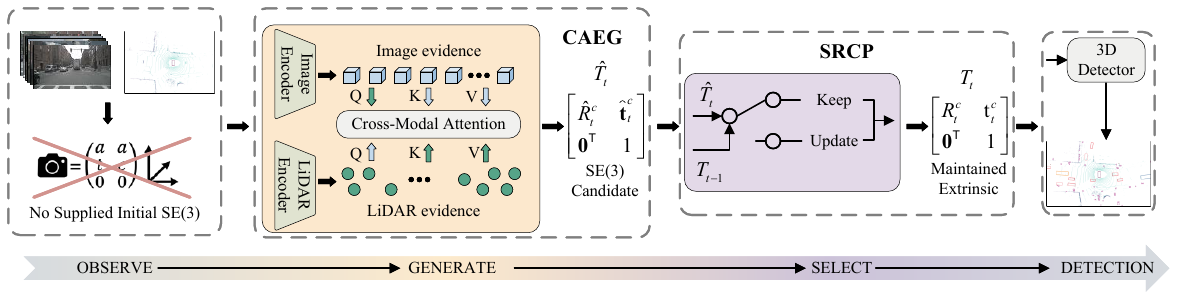}
    \caption{Overview of \method. \generator{} forms an absolute camera-to-LiDAR $\SE$ candidate $\widehat T_t$ from image--point evidence and coarse view context, without a supplied numerical initial transform. \policy{} selects \textsc{keep} or \textsc{update} between the previous working state $T_{t-1}$ and current candidate; the committed state $T_t$ supplies the frozen 3D detector. Camera superscripts are omitted.}
    \label{fig:overview}
\end{figure*}

The \emph{observe--generate--select--detect} loop in Fig.~\ref{fig:overview} begins with
\begin{equation}
 (\widehat T_t^c,g_t^c)=\mathcal{G}_{\theta}(I_t^c,P_t,e^c),
 \label{eq:recovery_interface}
\end{equation}
where $e^c$ is coarse camera-view context and $g_t^c$ is a cross-modal descriptor. The descriptor and observable quality/history statistics form policy observation $x_t^c$. A recurrent policy produces a commitment action, and (\ref{eq:switch}) selects the working transform. The detector then consumes $T_t^c$ for image lifting and fusion. This interface separates evidence-based recovery, state acceptance, and downstream consumption while keeping detector weights fixed.

\begin{figure*}[t]
    \centering
    \includegraphics[width=0.96\textwidth]{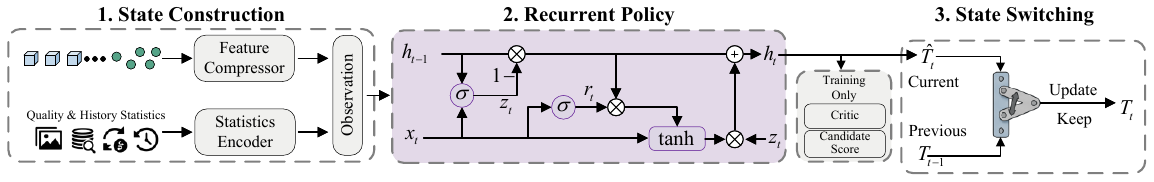}
    \caption{Detailed \policy{} architecture. The CAEG descriptor and quality/history statistics form $x_t$. The GRU memory update gate $z_t$ and reset gate $r_t$ update $h_t$; the actor's extrinsic commitment action $a_t$ selects $\widehat T_t$ or $T_{t-1}$ to establish $T_t$. The critic and candidate-error head provide training auxiliaries. Camera superscripts are omitted.}
    \label{fig:architecture}
\end{figure*}

\subsection{Cross-Modal Absolute Extrinsic Generator}
\label{subsec:caeg}

\subsubsection{Image and Point Representations}

The \generator{} block in Fig.~\ref{fig:overview} is implemented as follows. Each image is resized to $128\times352$ and processed by a four-stage shared convolutional encoder. Four stride-2 reductions followed by adaptive $4\times11$ pooling produce 44 image tokens per camera. A learned 44-token positional embedding $p_I$ and camera view token $e^c$ are then added,
\begin{equation}
    F_{I,t}^{c}=f_I(I_t^c)+p_I+\mathbf{1}_{44}(e^c)^{\mathsf T}
    \in\mathbb{R}^{44\times d}, \quad d=128.
\end{equation}
The shared encoder keeps the parameter count independent of camera count. Coarse view context combines nominal-yaw encoding $(\sin\psi_c,\cos\psi_c)$ with learned front/back/left/right embeddings. It supplies coarse mounting direction and view identity; the complete rotation and translation remain observation-conditioned outputs.

For LiDAR, $x$ and $y$ are normalized by 54 m, $z$ by 8 m, and reflectance by 255. A pointwise multilayer perceptron maps the four channels through widths 64, 128, and 128:
\begin{equation}
    F_{P,t}=f_P(P_t)\in\mathbb{R}^{N\times d},\qquad N=1024.
\end{equation}
No projection of LiDAR points into an image is used at this stage, because such a projection would already require the unknown transform.

\subsubsection{Projection-Free Bidirectional Feature Interaction}

The image and point tokens exchange evidence through two multihead-attention operations. For camera $c$,
\begin{align}
 \widetilde F_{I,t}^{c} &= \operatorname{LN}_{I}\!\left(F_{I,t}^{c}+\operatorname{MHA}(F_{I,t}^{c},F_{P,t},F_{P,t})\right), \\
 \widetilde F_{P,t}^{c} &= \operatorname{LN}_{P}\!\left(F_{P,t}+\operatorname{MHA}(F_{P,t},F_{I,t}^{c},F_{I,t}^{c})\right),
\end{align}
where the three arguments denote query, key, and value. The image-to-point branch asks which 3D observations explain each image token, whereas the point-to-image branch asks which appearance evidence supports each point. Since attention is formed in feature space, neither branch requires an extrinsic matrix.

The two streams are pooled by complementary operators and concatenated:
\begin{equation}
 g_t^c=\left[
 \operatorname{mean}(\widetilde F_{I,t}^{c});
 \operatorname{max}(\widetilde F_{I,t}^{c});
 \operatorname{max}(\widetilde F_{P,t}^{c})
 \right]\in\mathbb{R}^{3d}.
\end{equation}
A $384\!\rightarrow\!256\!\rightarrow\!128\!\rightarrow\!9$ head predicts translation $\widehat{\boldsymbol t}_t^c\in\mathbb{R}^{3}$ and a continuous 6-D rotation representation~\cite{zhou2019rotation}. Orthogonalization converts the latter to $\widehat R_t^c\in\SO$, yielding
\begin{equation}
 \widehat T_t^c=
 \begin{bmatrix}
  \widehat R_t^c & \widehat{\boldsymbol t}_t^c\\
  \boldsymbol{0}^{\mathsf T} & 1
 \end{bmatrix}.
\end{equation}
The resulting complete transform establishes the absolute candidate in (\ref{eq:recovery_interface}).

\subsubsection{Geometry-Regularized Supervision}

The generator is trained with five complementary terms:
\begin{equation}
 \mathcal{L}_{\rm gen}=\mathcal{L}_{t}+\mathcal{L}_{R}
 +\mathcal{L}_{6D}+0.2\mathcal{L}_{\rm anc}+0.1\mathcal{L}_{\rm att}.
 \label{eq:generator_loss}
\end{equation}
$\mathcal{L}_{t}$ is a Smooth-$L_1$ translation loss, $\mathcal{L}_{R}$ is the geodesic angle on $\SO$, and $\mathcal{L}_{6D}$ regularizes the continuous rotation coordinates. $\mathcal{L}_{\rm anc}$ applies both predicted and reference transforms to the origin, the three unit-axis points, and $(0,0,5)^{\mathsf T}$, coupling rotation and translation in metric space. For $\mathcal{L}_{\rm att}$, the training-only reference transform, intrinsics, and image augmentation project valid points onto the $4\times11$ token grid; a reverse-attention negative log-likelihood encourages spatially meaningful support. The loss couples pose-parameter accuracy, rigid-body geometric action, and cross-modal spatial correspondence, providing complementary supervision for complete $\SE$ recovery.

\subsection{Selective Recurrent Calibration Policy}
\label{subsec:srcp}

\subsubsection{Working-State Commitment}

The first candidate establishes the working geometry, $T_0^c=\widehat T_0^c$. Each subsequent action determines which extrinsic is used until another candidate is accepted:
\begin{equation}
 T_t^c=
 \begin{cases}
   T_{t-1}^c, & a_t^c=\textsc{keep},\\
   \widehat T_t^c, & a_t^c=\textsc{update}.
 \end{cases}
 \label{eq:switch}
\end{equation}
This persistence makes commitment a sequential decision. Retaining a reliable state filters candidate fluctuations, while accepting a reliable new candidate adapts the working geometry to a lasting change. \policy{} therefore evaluates the current evidence in the context of the state's acceptance history, as illustrated in Fig.~\ref{fig:architecture}.

\subsubsection{Observable State and Recurrent Memory}

The decision combines three complementary sources. \emph{Cross-modal evidence} is represented by $g_t^c$, compressed from 384 to 128 dimensions. \emph{Observation quality} comprises attention entropy and peak, image sharpness, LiDAR coverage, and cross-modal consistency. \emph{State history} comprises candidate-to-maintained and candidate-to-previous-candidate rotation/translation differences, history age, and the previous action. These statistics and a constant bias form a 12-D vector. Rotation, translation, and age are normalized by $30^{\circ}$, 2 m, and 40 frames, respectively; dynamic ratios are clipped at 4. A 64-D statistic encoding is concatenated with the descriptor and projected to the 128-D observation $x_t^c$. Quality describes the support for a candidate, while history describes the persistence and consistency of its geometric change.

A shared 128-D GRU maintains separate temporal memory for each camera:
\begin{equation}
 h_t^c=\operatorname{GRUCell}(x_t^c,h_{t-1}^c).
 \label{eq:recurrent_policy}
\end{equation}
In Fig.~\ref{fig:architecture}, $z_t^c$ is the GRU memory update gate (shown in the memory-write convention), and $r_t^c$ is the reset gate. These gates update internal memory; the actor's action $a_t^c$ separately commits the working extrinsic in (\ref{eq:switch}). The hidden state also feeds a scalar critic and nonnegative candidate-error head. Shared weights and camera-specific memory support the same maintenance rule across views.

\subsubsection{Sequence-Level Learning Objective}

The actor samples actions during training and selects the most likely action during inference. Learning rewards the quality of the working state produced by those decisions.

For policy training, pose quality is summarized by
\begin{equation}
 E_t(T)=\frac{\Delta R_{\rm deg}(T,T_t^*)}{5}
      +\frac{\Delta t_{\rm m}(T,T_t^*)}{0.25},
 \label{eq:composite_error}
\end{equation}
where $T_t^*$ supplies the training reference for the reward and auxiliary candidate score. The immediate reward is
\begin{align}
 r_t={}&E_t(T_{t-1})-E_t(T_t)
 -2\max\{0,E_t(T_t)-E_t(T_{t-1})\} \nonumber\\
 &-0.02\,\mathbb{I}[a_t=\textsc{update}],
 \label{eq:reward}
\end{align}
and the terminal reward adds
\begin{equation}
 r_{\rm term}=-2E_T(T_T)-0.5\max_t E_t(T_t).
 \label{eq:terminal_reward}
\end{equation}
The reward comprises \emph{state improvement}, an asymmetric \emph{harmful-commitment penalty}, and a \emph{commitment cost}. These terms favor beneficial candidates, penalize worsening state changes more strongly, and discourage unproductive commitments. The terminal reward additionally optimizes final and sequence-peak error. PPO learns present decisions through their continuing effect on the working geometry.

\subsubsection{PPO Optimization}

PPO~\cite{schulman2017ppo} uses generalized advantage estimation ($\gamma=0.99$, $\lambda=0.95$), clipping ratio 0.2, and four update epochs per rollout:
\begin{equation}
 \mathcal{L}_{\rm policy}=\mathcal{L}_{\rm clip}
 +0.5\mathcal{L}_{\rm value}-0.02\mathcal{H}
 +0.2\mathcal{L}_{\rm score},
 \label{eq:ppo_loss}
\end{equation}
Here $\mathcal{L}_{\rm value}=\tfrac{1}{2}\operatorname{MSE}$, $\mathcal{H}$ is action entropy, and $\mathcal{L}_{\rm score}$ supplies Smooth-$L_1$ auxiliary supervision for candidate quality. Actor/value terms exclude forced initialization; score supervision covers valid frames. At inference, the actor decides commitment from observations and maintained history.

\subsection{Frozen-Detector Interface and Complexity}

At inference, the maintained $T_t^c$ supplies the camera-to-LiDAR geometry used by image lifting and cross-modal fusion. The detector checkpoint and preprocessing remain fixed across extrinsic sources, making geometric recovery and commitment the adaptation interface. Candidate generation runs on each frame; the commitment rate describes writes to the working state. \generator{} has 629,673 trainable parameters and \policy{} has 179,076, totaling 808,749 parameters outside the frozen detector.

\section{Experiments}
\label{sec:experiments}

\begin{figure*}[t]
 \centering
 \includegraphics[width=0.98\textwidth]{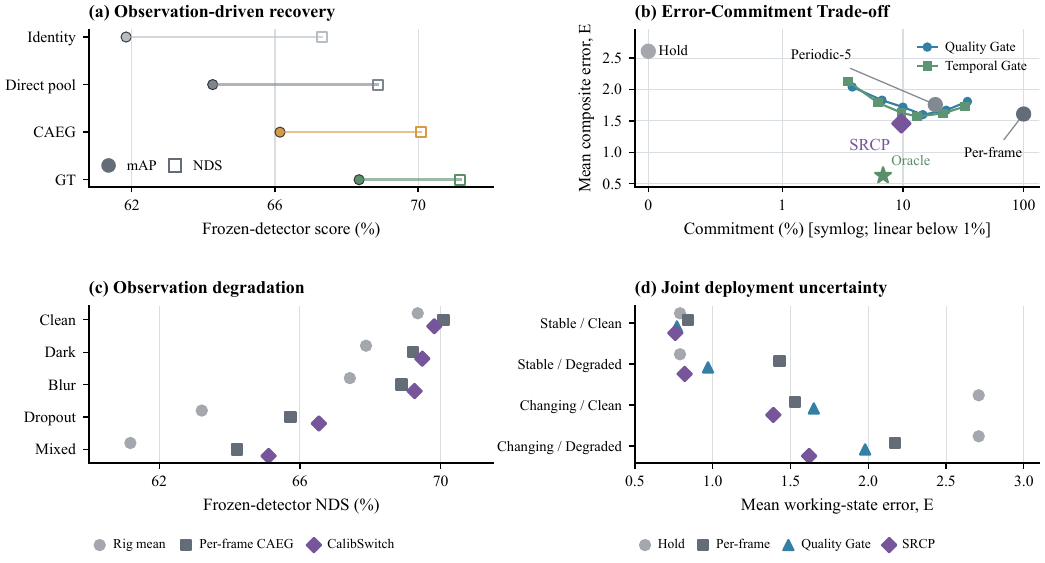}
 \caption{Recovery and maintenance. (a) Frozen detection with four extrinsic sources. (b) Gate threshold sweeps and reference policies; the commitment axis uses symmetric-log scaling with a linear 0--1\% region to display zero-commitment Hold. Lines connect tested thresholds. (c) Full-sequence corruption. (d) Paired joint geometry--observation uncertainty.}
 \label{fig:results_summary}
\end{figure*}

The evaluation follows candidate recovery, selective maintenance, and perception, connected in Fig.~\ref{fig:results_summary}.

\subsection{Experimental Setup}

\subsubsection{Data and Geometric Protocol}

We train on the official nuScenes training split and evaluate on the chronological 150-scene, 6,019-frame validation split~\cite{caesar2020nuscenes}, using all six cameras and 1024 LiDAR points per generator sample. Training samples rotation components within $\pm20^{\circ}$ and translation within $\pm(1.5,1.5,1.0)$ m, including 10\% identity samples. Table~\ref{tab:pose_comparison} reports the common $\alpha=1.25$ condition: $\pm25^{\circ}$ and $\pm(1.875,1.875,1.25)$ m.
Table~\ref{tab:ood_range_generalization} scales both bounds by $\alpha=1.00$, $1.25$, and $1.50$ with matched configurations across methods.

Coordinate-consistent virtual geometric reconfiguration provides controlled state evaluation: for sampled $S\in\SE$, points become $\boldsymbol p_i'=R_S\boldsymbol p_i+\boldsymbol t_S$, reflectance is preserved, and targets satisfy $T^{c*'}=ST^{c*}$. Observations and targets share LiDAR coordinates. For detection at each time $t$, the maintained state is converted to the recorded frame through $S_t^{-1}T_t^{c'}$; detector points, annotations, intrinsics, and image augmentation retain their recorded convention for all sources. Supplied-transform baselines receive their prescribed initialization; \generator{} receives images, points, and coarse view context. Natural real-platform sequences complement this controlled evaluation with actual installation and operating observations.

\begin{table*}[t]
\caption{Absolute Extrinsic Accuracy Under Virtual Rig Reconfiguration. Methods are grouped by transformation-matrix input; lower is better.}
\label{tab:pose_comparison}
\centering
\footnotesize
\setlength{\tabcolsep}{8pt}
\begin{tabular}{lccccc}
\toprule
Method & Numerical initial $T$ & $t_{\rm mean}$ (m)$\downarrow$ & $t_{\rm P95}$ (m)$\downarrow$ & $R_{\rm mean}$ (deg)$\downarrow$ & $R_{\rm P95}$ (deg)$\downarrow$\\
\midrule
\multicolumn{6}{l}{\emph{With numerical initial transform}}\\
Training-rig mean & Yes & 1.725 & 2.566 & 23.997 & 34.493\\
CalibNet~\cite{iyer2018calibnet} & Yes & 0.423 & 1.110 & 9.180 & 24.40\\
LCCNet~\cite{lv2021lccnet} & Yes & 0.351 & 0.940 & 7.760 & 21.10\\
CalibFormer~\cite{xiao2024calibformer} & Yes & 0.284 & 0.780 & 6.440 & 17.80\\
\addlinespace[2pt]
\multicolumn{6}{l}{\emph{Without a numerical transformation matrix}}\\
Identity-$\SE$ diagnostic & No & 1.840 & 2.780 & 102.60 & 124.80\\
Direct pooled regressor & No & 0.338 & 0.890 & 7.470 & 20.20\\
\generator{} (ours) & \textbf{No} & \textbf{0.248} & \textbf{0.671} & \textbf{5.884} & \textbf{15.72}\\
\bottomrule
\end{tabular}
\end{table*}

\begin{table}[tb]
\caption{Mean Pose Error Beyond the Training Perturbation Range. Each entry is $t$ (m)/$R$ (deg); $\alpha>1$ is OOD.}
\label{tab:ood_range_generalization}
\centering
\footnotesize
\setlength{\tabcolsep}{1.3pt}
\begin{tabular}{@{}lcccc@{}}
\toprule
Method & Init. $T$ & $\alpha=1.00$ & $\alpha=1.25$ & $\alpha=1.50$\\
\midrule
CalibNet~\cite{iyer2018calibnet} & Yes & 0.283/6.12 & 0.423/9.18 & 0.782/17.43\\
LCCNet~\cite{lv2021lccnet} & Yes & 0.231/5.11 & 0.351/7.76 & 0.638/14.26\\
CalibFormer~\cite{xiao2024calibformer} & Yes & 0.190/4.30 & 0.284/6.44 & 0.517/11.91\\
\addlinespace[2pt]
Direct pooled regressor & No & 0.251/5.57 & 0.338/7.47 & 0.471/10.63\\
\generator{} (ours) & \textbf{No} & \textbf{0.183/4.16} & \textbf{0.248/5.88} & \textbf{0.341/7.96}\\
\bottomrule
\end{tabular}
\end{table}

\subsubsection{Continuous Operation and Comparators}

\emph{Protocol I: Sustained Extrinsic-State Transition} preserves scene order. Two thirds of sequences undergo a persistent geometric change, using rotation bounds of $\pm15^{\circ}$ and translation bounds of $\pm(1.0,1.0,0.75)$ m; the remainder retain their geometry.
One change onset is sampled within zero-based frames 12--23, and the new geometry remains until the scene ends. Protocol I supplies Table~\ref{tab:temporal_policy} and Fig.~\ref{fig:results_summary}(b). Separate full-sequence darkness, blur, point dropout, and mixed corruption supply Fig.~\ref{fig:results_summary}(c).

\emph{Protocol II: Randomized Joint-Uncertainty Operation} crosses stable/changing geometry with clean/degraded observations (Table~\ref{tab:joint_uncertainty}, Fig.~\ref{fig:maintenance_dynamics}). Each pair shares scene, camera, geometry trajectory, seed, and initial candidate/working state. Changes start after six frames, leave at least eight frames, and persist to scene end. Degradation affects only finite windows sampled independently of geometric onset, balancing darkness, blur, point dropout, and mixed cases. Policies share each condition's CAEG cache, isolating observation reliability from geometry.

We retain First-frame Hold, Periodic-5, Per-frame Update, and Oracle Gain as operating references. Quality Gate thresholds a frame-wise score fitted from current quality statistics to reference candidate error. Temporal Gate tests consecutive-pose consistency with hysteresis. Thresholds are selected on training-split development scenes for commitments near the SRCP rate, then fixed for validation. Sweeps characterize the error--commitment trade-off on the same Protocol-I candidate cache in Fig.~\ref{fig:results_summary}(b). Each policy starts from the same first candidate.

\begin{table*}[t]
 \centering
\captionof{table}{Working-state maintenance in Protocol I with a shared candidate stream and frozen detector. Harmful rate is the fraction of worsening commitments; ``--'' denotes no commitments. Green annotations indicate gains over per-frame updating; mAP/NDS gains are in points.}
\label{tab:temporal_policy}
\footnotesize
\setlength{\tabcolsep}{3.0pt}
\begin{tabularx}{\textwidth}{lG{0.70}G{1.10}G{1.15}G{1.05}G{0.80}G{0.85}G{1.175}G{1.175}}
\toprule
Policy & Commits & \shortstack[c]{Commit\\(\%) $\downarrow$} & \shortstack[c]{Harmful\\(\%) $\downarrow$} & Mean $E$ $\downarrow$ & Final $E$ $\downarrow$ & \shortstack[c]{Mean Peak\\$E$ $\downarrow$} & mAP $\uparrow$ & NDS $\uparrow$\\
\midrule
First-frame hold & 0 & 0.00 & -- & 2.61 & 3.98 & 4.05 & 62.42 & 67.87\\
Periodic-5 & 6522 & 18.52 & 48.10 & 1.76 & 2.49 & 4.17 & 64.15 & 68.94\\
Per-frame update & 35214 & 100.00 & 49.20 & 1.61 & 2.34 & 3.76 & 64.61 & 69.19\\
Quality Gate & 3518 & 9.99 & 3.24 & 1.72 & 2.13 & 2.84 & 65.04 & 69.41\\
Temporal Gate & 3429 & 9.74 & 2.22 & 1.63 & 2.07 & 2.78 & 65.12 & 69.56\\
\rowcolor{OursTint}
\gaincell{\policy{} (ours)}{} & \gaincell{3415}{} & \gaincell{9.70}{\gdown{90.3}} & \gaincell{0.54}{\gdown{98.9}} & \gaincell{1.46}{\gdown{9.3}} & \gaincell{1.86}{} & \gaincell{2.41}{} & \gaincell{66.04}{\gup{1.43}} & \gaincell{70.11}{\gup{0.92}}\\
Oracle gain & 2409 & 6.84 & 0.00 & 0.63 & 0.53 & 1.67 & 66.82 & 70.48\\
\bottomrule
\end{tabularx}

\end{table*}
\begin{table*}[t]
\caption{Paired joint uncertainty in Protocol II. Clean/degraded pairs share geometry and initialization. $D_{\mathrm{rec}}$: recovery delay (frames); harmful rate is conditional on commitment; ``--'': undefined or inapplicable. Green annotations are relative to per-frame updating within each condition.}
\label{tab:joint_uncertainty}
\centering\footnotesize
\setlength{\tabcolsep}{4pt}
\begin{tabularx}{\textwidth}{llG{0.75}G{1.20}G{0.75}G{0.80}G{1.15}G{1.35}}
\toprule
Condition & Policy & Mean $E$ $\downarrow$ & \shortstack[c]{Mean Peak\\$E$ $\downarrow$} & \shortstack[c]{Commit\\(\%) $\downarrow$} & \shortstack[c]{Harmful\\(\%) $\downarrow$} & $D_{\mathrm{rec}}$ $\downarrow$ & NDS $\uparrow$\\
\midrule
Stable / Clean & Hold & 0.79 & 0.79 & 0.00 & -- & -- & 69.72\\
 & Per-frame & 0.84 & 2.16 & 100.00 & 47.83 & -- & 70.06\\
 & Quality Gate & 0.77 & 1.02 & 7.36 & 3.12 & -- & 69.98\\
\rowcolor{OursTint}
 & \textbf{\policy{} (ours)} & \textbf{0.76} & \textbf{0.91} & \textbf{1.83} & \textbf{0.63} & -- & \textbf{70.04}\\
\addlinespace[2pt]
Stable / Degraded & Hold & 0.79 & 0.79 & 0.00 & -- & -- & 68.11\\
 & Per-frame & 1.43 & 4.86 & 100.00 & 53.77 & -- & 66.92\\
 & Quality Gate & 0.97 & 1.19 & 8.41 & 4.69 & -- & 68.84\\
\rowcolor{OursTint}
 & \gaincell{\policy{} (ours)}{} & \gaincell{0.82}{} & \gaincell{0.94}{\gdown{80.7}} & \gaincell{2.37}{} & \gaincell{0.82}{} & \gaincell{--}{} & \gaincell{69.31}{\gup{2.39}}\\
\addlinespace[2pt]
Changing / Clean & Hold & 2.71 & 4.62 & 0.00 & -- & -- & 67.03\\
 & Per-frame & 1.53 & 3.43 & 100.00 & 47.62 & 2.40 & 69.44\\
 & Quality Gate & 1.65 & 3.71 & 10.08 & 3.45 & 4.80 & 69.57\\
\rowcolor{OursTint}
 & \gaincell{\policy{} (ours)}{} & \gaincell{1.39}{} & \gaincell{2.38}{\gdown{30.6}} & \gaincell{9.42}{} & \gaincell{0.48}{} & \gaincell{2.90}{} & \gaincell{70.29}{\gup{0.85}}\\
\addlinespace[2pt]
Changing / Degraded & Hold & 2.71 & 4.62 & 0.00 & -- & -- & 64.28\\
 & Per-frame & 2.17 & 6.42 & 100.00 & 54.86 & 5.80 & 66.53\\
 & Quality Gate & 1.98 & 4.79 & 10.77 & 5.38 & 7.40 & 67.66\\
\rowcolor{OursTint}
 & \gaincell{\policy{} (ours)}{} & \gaincell{1.62}{} & \gaincell{3.18}{\gdown{50.5}} & \gaincell{10.11}{} & \gaincell{0.71}{} & \gaincell{4.10}{\gdown{29.3}} & \gaincell{68.57}{\gup{2.04}}\\
\bottomrule
\end{tabularx}
\end{table*}

\subsubsection{Training and Evaluation Metrics}

\generator{} uses eight epochs of AdamW with learning rate $3\times10^{-4}$, weight decay $10^{-2}$, cosine decay, bfloat16 training, batch size 32 per GPU, and gradient clipping at 20. \policy{} uses 12 epochs, episode batches of 25, AdamW at $3\times10^{-4}$, weight decay $10^{-2}$, and clipping at 5. Detector weights, preprocessing, checkpoint, and evaluation settings remain fixed throughout each comparison.

\begin{figure*}[t]
 \centering
 \includegraphics[width=0.96\textwidth]{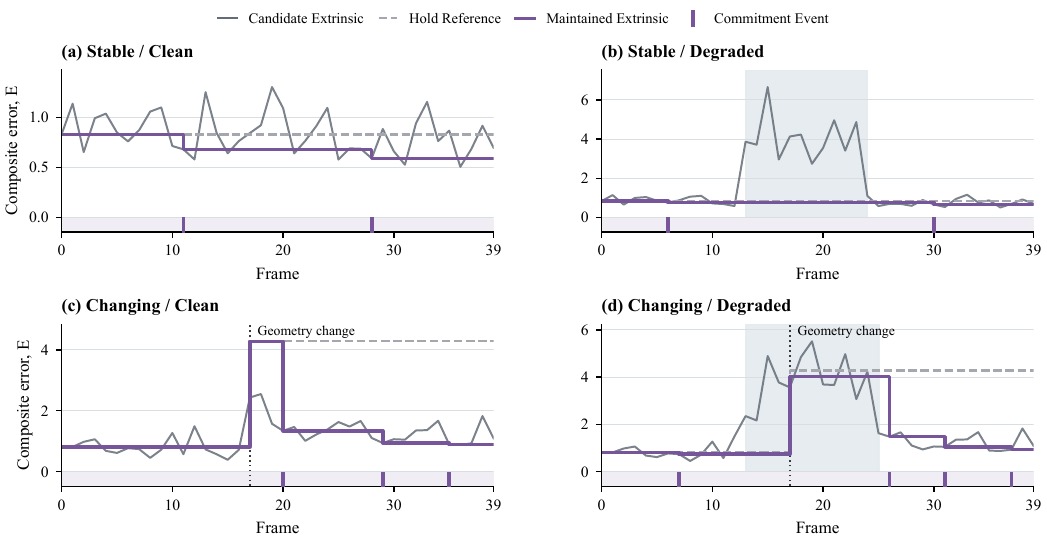}
 \caption{Candidate estimates and working geometry in paired chronological examples. Clean/degraded panels share geometry and initialization within each row. Curves show candidate, Hold-reference, and maintained-extrinsic error. Shading denotes observation degradation, dotted lines mark persistent geometric changes, and purple ticks in the lower action strip mark commitments. A reference change alters held-state error. Panels use different error ranges to resolve each condition.}
 \label{fig:maintenance_dynamics}
\end{figure*}

Recovery uses mean and P95 rotation/translation errors; perception uses official mAP, NDS, mATE, and mAOE. Maintenance uses mean and final composite error in (\ref{eq:composite_error}), together with mean peak error.
For $M$ camera-level sequences $\mathcal S_j$, mean peak error is
{
\begin{equation}
 E_{\rm peak}=\frac{1}{M}\sum_{j=1}^{M}\max_{t\in\mathcal S_j}E_{j,t}.
 \label{eq:mean_peak}
\end{equation}
}
Commitment rate divides accepted post-initialization candidates by valid camera decisions (35,214 in Protocol I). Harmful rate is the fraction of commitments that worsen current reference error; it is undefined for Hold and shown as ``--''.
Recovery delay $D_{\rm rec}$ counts frames from geometric onset to the first of three consecutive frames with $E\leq1.5$, capped by the remaining episode length. It is averaged over all change events for adaptive policies. The four chronological examples in Fig.~\ref{fig:maintenance_dynamics} explain behavior alongside full-scene aggregates.

\subsection{Observation-Driven Extrinsic Recovery}

Tables~\ref{tab:pose_comparison} and~\ref{tab:ood_range_generalization} establish the candidate stream needed for online maintenance. \generator{} obtains mean errors of 0.248 m/5.884$^\circ$ and P95 errors of 0.671 m/15.72$^\circ$, recovering complete geometry without a numerical initial matrix. Relative to independent pooled regression, mean translation and rotation errors improve by 26.6\% and 21.2\%. Bidirectional image--point evidence therefore contributes to the recovery of usable absolute candidates.

The range-generalization comparison keeps training support fixed for all methods. At $\alpha=1.50$, \generator{} reaches 0.341 m/7.96$^\circ$, compared with 0.517 m/11.91$^\circ$ for CalibFormer and 0.471 m/10.63$^\circ$ for pooled absolute regression. These results establish CAEG as an observation-driven source of complete extrinsic candidates across geometric configurations. The following experiments examine their selective commitment to reliable working geometry during continuous operation.

\begin{figure*}[t]
 \centering
 \includegraphics[width=0.98\textwidth]{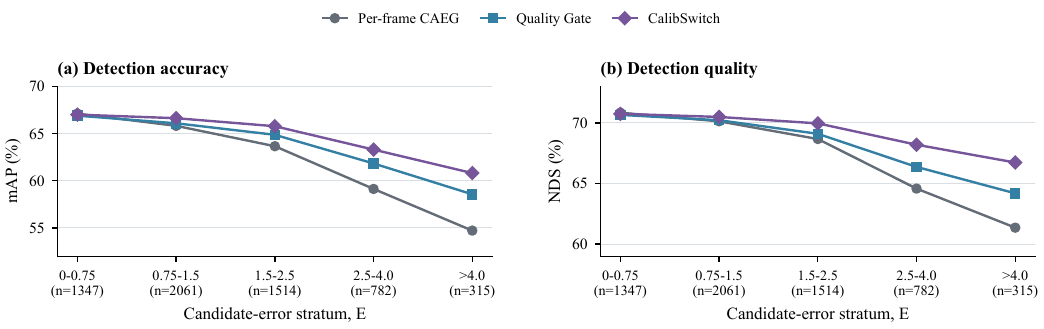}
 \caption{Geometric reliability and frozen perception on five common Protocol-I candidate-error strata. (a) mAP and (b) NDS for per-frame CAEG, Quality Gate, and CalibSwitch. Each stratum uses the same frames across policies; $n$ denotes its frame count. Lines connect discrete strata. Candidate error defines grouping, while the detector consumes each policy's working geometry.}
 \label{fig:geometry_perception}
\end{figure*}

\begin{table}[tb]
\caption{Frozen-detector results with alternative extrinsic sources. Bold marks the best method without a numerical initial transform.}
\label{tab:detection_main}
\centering
\footnotesize
\setlength{\tabcolsep}{1.0pt}
\begin{tabularx}{\columnwidth}{@{}>{\raggedright\arraybackslash}Xccccc@{}}
\toprule
Source & Init. $T$ & mAP$\uparrow$ & NDS$\uparrow$ & mATE$\downarrow$ & mAOE$\downarrow$\\
\midrule
GT extrinsics & GT & 68.35 & 71.18 & 0.2792 & 0.2990\\
Rig nominal & Yes & 68.14 & 71.01 & 0.2798 & 0.3003\\
CalibFormer & Yes & 65.29 & 69.62 & 0.2855 & 0.3081\\
Pooled regressor & No & 64.26 & 68.87 & 0.2889 & 0.3146\\
LiDAR-only & -- & 62.20 & 67.62 & 0.2924 & 0.3248\\
Identity $\SE$ & No & 61.84 & 67.31 & 0.2941 & 0.3304\\
\midrule
\generator{} (ours) & \textbf{No} & \textbf{66.14} & \textbf{70.08} & \textbf{0.2834} & \textbf{0.3047}\\
\bottomrule
\end{tabularx}
\end{table}

\subsection{Selective Extrinsic-State Maintenance}

Table~\ref{tab:temporal_policy} evaluates state commitment with a shared CAEG candidate stream and frozen detector. Relative to per-frame updating, SRCP reduces the commitment rate by 90.3\% (100.00\% to 9.70\%), the harmful-commitment fraction by 98.9\% (49.20\% to 0.54\%), and mean working-state error by 9.3\% (1.61 to 1.46). Relative to first-frame holding, mean and mean peak errors decrease by 44.1\% and 40.5\%.

At comparable commitment budgets (9.70--9.99\%), SRCP lowers mean error by 15.1\% and 10.4\% relative to Quality Gate and Temporal Gate. Figure~\ref{fig:results_summary}(b) links this advantage to joint observation reliability and recurrent geometric history.

Table~\ref{tab:joint_uncertainty} evaluates paired geometric and observation uncertainty in Protocol II. Relative to per-frame updating, SRCP reduces mean peak error by 80.7\% (4.86 to 0.94) under Stable/Degraded operation and by 30.6\% (3.43 to 2.38) under Changing/Clean operation. When geometric changes coincide with degradation, it reduces mean peak error by 50.5\% (6.42 to 3.18) and recovery delay by 29.3\% (5.80 to 4.10 frames). Selective commitment thus preserves reliable geometry during transient disturbances while adapting to persistent changes.

Figure~\ref{fig:maintenance_dynamics} explains the decisions behind the aggregates. Temporary degradation raises candidate error while the accepted state remains stable. Following a persistent change, supported candidates establish new working geometry; when change and degradation overlap, commitment follows the return of informative observations. The paired trajectories distinguish candidate disturbances from reference-state changes. The mechanism analysis in Section~\ref{subsec:mechanism_analysis} further links these behaviors to quality awareness, recurrent memory, and the commitment objective.

\begin{table*}[t]
\centering
\begin{minipage}[t]{0.48\textwidth}
\captionof{table}{CAEG ablation under virtual rig reconfiguration. The detector and training schedule are fixed.}
\label{tab:caeg_ablation}
\centering
\footnotesize
\setlength{\tabcolsep}{2.0pt}
\begin{tabularx}{\columnwidth}{@{}>{\raggedright\arraybackslash}Xcccc@{}}
\toprule
Variant & $\bar t$ (m)$\downarrow$ & $\bar R$ (deg)$\downarrow$ & mAP$\uparrow$ & NDS$\uparrow$\\
\midrule
w/o view prior & 0.356 & 8.180 & 64.31 & 68.91\\
Pooling, no cross-attn. & 0.338 & 7.590 & 64.66 & 69.08\\
One-way cross-attn. & 0.292 & 6.570 & 65.28 & 69.49\\
w/o anchor loss & 0.279 & 6.310 & 65.49 & 69.62\\
w/o reverse-proj. loss & 0.267 & 6.150 & 65.67 & 69.73\\
w/o both geom. losses & 0.309 & 7.010 & 64.93 & 69.27\\
\midrule
Full \generator{} & \textbf{0.248} & \textbf{5.884} & \textbf{66.14} & \textbf{70.08}\\
\bottomrule
\end{tabularx}

\end{minipage}\hfill
\begin{minipage}[t]{0.48\textwidth}
\captionof{table}{SRCP mechanism analysis with the same CAEG candidate stream.}
\label{tab:srcp_ablation}
\centering\footnotesize
\setlength{\tabcolsep}{3pt}
\begin{tabularx}{\columnwidth}{@{}>{\raggedright\arraybackslash}Xrrrr@{}}
\toprule
Variant & Mean $E$ & \shortstack{Mean Peak\\$E$} & \shortstack{Harmful\\(\%)} & NDS\\
\midrule
w/o quality statistics & 1.75 & 3.08 & 4.73 & 69.38\\
w/o recurrent memory & 1.68 & 2.87 & 2.84 & 69.61\\
w/o asymmetric penalty & 1.53 & 2.94 & 3.96 & 69.76\\
\midrule
Full \policy{} & \textbf{1.46} & \textbf{2.41} & \textbf{0.54} & \textbf{70.11}\\
\bottomrule
\end{tabularx}

\end{minipage}
\end{table*}

\begin{figure*}[t]
 \centering
 \subfloat[Compact UGV.]{\includegraphics[width=0.30\textwidth]{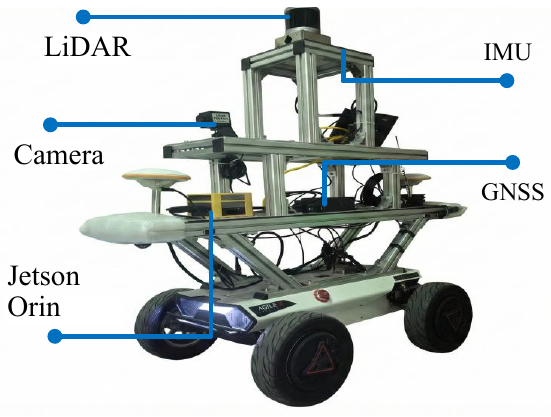}}\hspace{0.04\textwidth}
 \subfloat[Road vehicle.]{\includegraphics[width=0.30\textwidth]{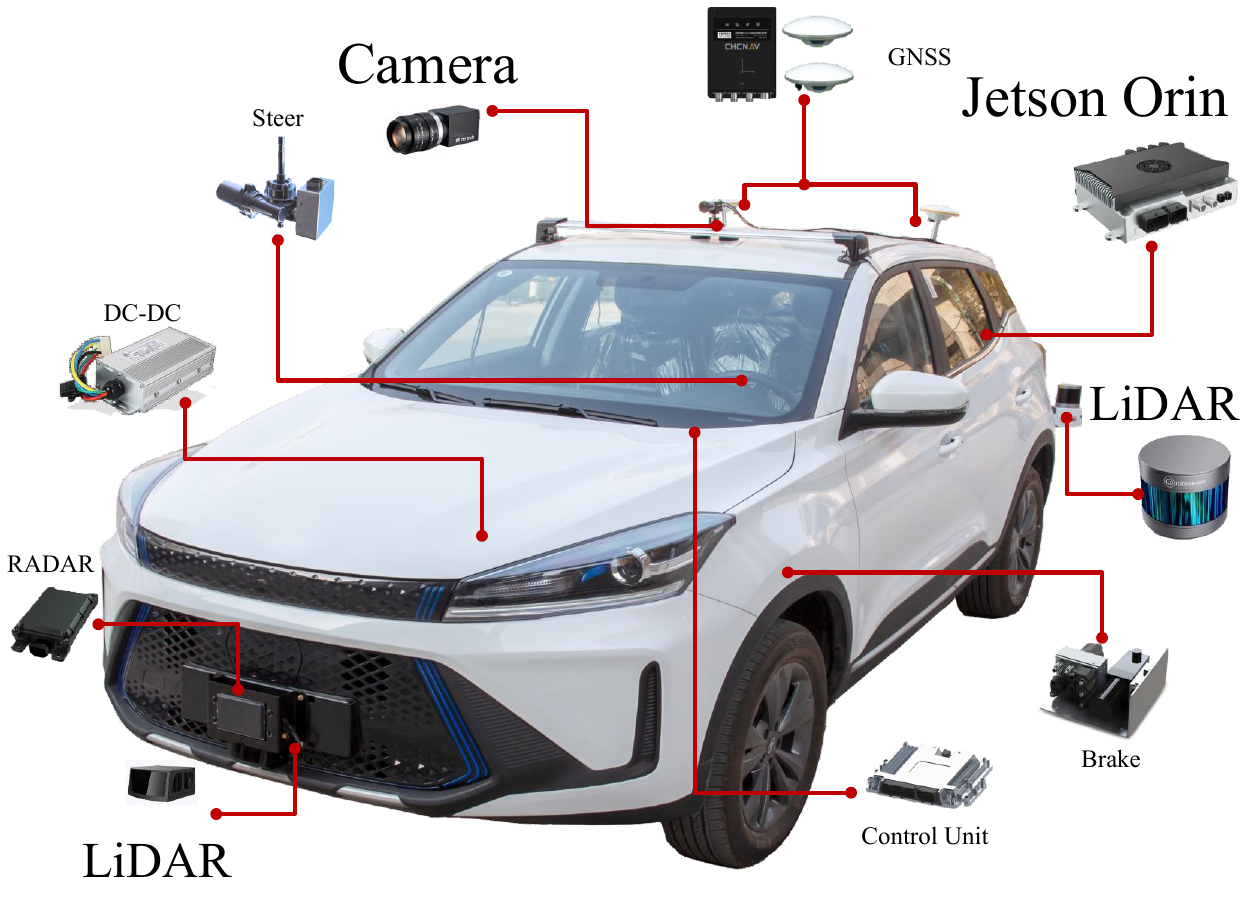}}
 \caption{Real-vehicle platforms. Both use Jetson Orin; this work evaluates their camera--LiDAR subset.}
 \label{fig:real_vehicle_platforms}
\end{figure*}

\begin{figure*}[t]
 \centering
 \includegraphics[width=0.92\textwidth]{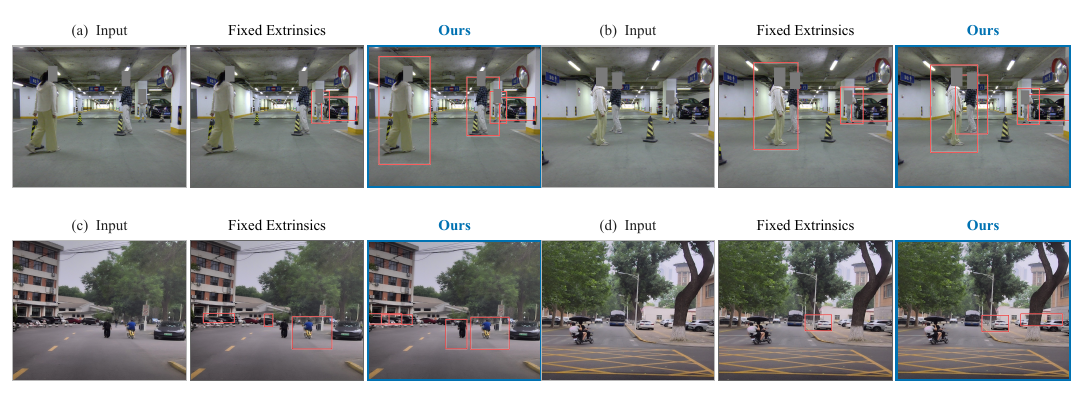}
 \caption{Natural two-platform deployment. Panels (a)--(b) show UGV garage frames; (c)--(d) show cross-date road scenes. Fixed Extrinsics uses the original calibration-file transform unchanged throughout the sequence; Ours uses CalibSwitch. Input denotes the raw view. All branches share the field of view and privacy masks.}
 \label{fig:real_vehicle_qualitative}
\end{figure*}

\begin{table*}[t]
\caption{Working-state continuity and perception in natural deployment. Pose errors use the road-platform reference.}
\label{tab:natural_deployment}
\centering\footnotesize
\setlength{\tabcolsep}{4pt}
\begin{tabular*}{\textwidth}{@{\extracolsep{\fill}}llrrrrrr@{}}
\toprule
Platform & Policy & Commit (\%) & $\overline{\delta R}$ (deg) & $\overline{\delta t}$ (m) & TCS & $\bar t$ (m) & $\bar R$ (deg)\\
\midrule
I: UGV & Hold & 0.00 & 0.000 & 0.000 & 0.932 & -- & --\\
I: UGV & Per-frame & 100.00 & 2.810 & 0.184 & 0.918 & -- & --\\
I: UGV & CalibSwitch & 1.47 & 0.043 & 0.006 & 0.949 & -- & --\\
II: Road & Hold & 0.00 & 0.000 & 0.000 & 0.903 & 0.312 & 4.906\\
II: Road & Per-frame & 100.00 & 1.740 & 0.076 & 0.917 & 0.266 & 4.349\\
II: Road & CalibSwitch & 2.18 & 0.061 & 0.004 & 0.938 & 0.263 & 3.813\\
\bottomrule
\end{tabular*}
\end{table*}

\begin{table}[tb]
 \centering
\captionof{table}{Platform-I Jetson AGX Orin execution (latency: mean/P95; TCS: temporal consistency).}
\label{tab:jetson_execution}
\footnotesize
\setlength{\tabcolsep}{2.2pt}
\begin{tabularx}{\columnwidth}{@{}>{\raggedright\arraybackslash}Xccccc@{}}
\toprule
Scene & FPS$\uparrow$ & E2E (ms)$\downarrow$ & \generator{} (ms)$\downarrow$ & Commit (\%) & TCS$\uparrow$\\
\midrule
Hallway & 13.8 & 72.5/80.4 & 7.8/8.6 & 0.67 & 0.944\\
Garage & 13.3 & 75.2/84.7 & 8.2/9.1 & 1.86 & 0.958\\
Stair & 13.7 & 73.1/81.6 & 7.9/8.8 & 1.34 & 0.931\\
\midrule
\textbf{Overall} & \textbf{13.5} & \textbf{74.1/83.9} & \textbf{8.0/9.0} & \textbf{1.47} & \textbf{0.949}\\
\bottomrule
\end{tabularx}

\end{table}

\subsection{Impact on Multimodal Perception}

Table~\ref{tab:detection_main} connects initial recovery to the frozen detector. CAEG reaches 66.14\% mAP and 70.08\% NDS, improving over identity geometry by 4.30 and 2.77 points. It also improves over the pooled regressor by 1.88 mAP and 1.21 NDS points. This comparison holds the detection network fixed and exchanges its geometric input, directly linking observation-driven recovery to useful multimodal fusion.

The improved working-state reliability directly benefits frozen multimodal detection. In Table~\ref{tab:temporal_policy}, SRCP increases mAP from 64.61\% to 66.04\% (+1.43 points) and NDS from 69.19\% to 70.11\% (+0.92 points) over per-frame updating. Under Protocol II in Table~\ref{tab:joint_uncertainty}, the corresponding NDS gains are 2.39, 0.85, and 2.04 points for Stable/Degraded, Changing/Clean, and Changing/Degraded conditions, respectively. These gains connect selective extrinsic-state maintenance to more reliable downstream perception across deployment conditions.

Figure~\ref{fig:results_summary}(c) further shows that maintained geometry improves average NDS under corruption by 2.23 points over the fixed rig mean and 0.42 points over per-frame generation. Figure~\ref{fig:geometry_perception} groups the same 6,019 Protocol-I validation frames by frame-wise mean candidate $E$ across six cameras. Policies share masks, and each stratum uses the same official evaluation. Performance is close at low error; at $E>4.0$, CalibSwitch exceeds per-frame CAEG by \highErrorNDSGain{} NDS points. Its perception advantage becomes more pronounced as candidate reliability decreases.

\subsection{Mechanism Analysis}
\label{subsec:mechanism_analysis}

Table~\ref{tab:caeg_ablation} explains how observations become useful candidates. Coarse view context supplies mounting direction; removing it yields the largest pose degradation. Independent pooling loses 1.48 mAP points, and one-way interaction remains below the bidirectional design. Anchor-transform and reverse-projection supervision improve pose and detection quality. Bidirectional evidence and geometric supervision thus jointly support complete candidate recovery.

Table~\ref{tab:srcp_ablation} explains how candidates become reliable working states. Removing quality statistics raises harmful commitments to 4.73\%; removing recurrent memory raises mean $E$ to 1.68 while retaining explicit history statistics. Removing the asymmetric penalty increases harmful commitments and mean peak error. Quality awareness, recurrent memory, and asymmetric commitment penalties jointly improve maintained geometry. Candidate recovery and state commitment therefore provide complementary mechanisms in the same perception chain.

\subsection{Real-Vehicle and Embedded Validation}

Figures~\ref{fig:real_vehicle_platforms} and~\ref{fig:real_vehicle_qualitative} show natural deployment. Platform I provides single-camera UGV operation across 16 indoor/semistructured sequences and 2,400 frames. Platform II provides three-camera road data: 20 sequences and 4,785 frames, including 13 later-date test sequences and 3,203 frames. Figure~\ref{fig:real_vehicle_qualitative} compares the original fixed-calibration configuration with CalibSwitch; Table~\ref{tab:natural_deployment} compares state-use policies sharing CAEG candidates.

One-epoch road-platform generator adaptation yields 0.266 m/4.349$^\circ$; maintained geometry reaches 0.263 m/3.813$^\circ$. Adaptation uses seven earlier-date sequences (1,582 frames), platform extrinsic supervision, and the generator loss. The cross-date test is held out, and detector weights remain frozen.

Temporal detection consistency (TCS) measures adjacent-frame object matching. Same-class detections are associated one-to-one after ego-motion compensation with a fixed 2-m center-distance gate. For detection sets $D_{t-1},D_t$ and matches $M_t$,
{
\begin{equation}
 {\rm TCS}=\frac{1}{|\mathcal{A}|}\sum_{t\in\mathcal{A}}
 \frac{2|M_t|}{|D_{t-1}|+|D_t|},
 \label{eq:tcs}
\end{equation}
}
where $\mathcal{A}$ excludes double-empty pairs and sequence boundaries. All policies share score thresholds and matching rules. Adjacent-state variations are $\delta R_t=\angle(R_{t-1}^{\mathsf T}R_t)$ and $\delta t_t=\|\boldsymbol t_t-\boldsymbol t_{t-1}\|_2$.

Table~\ref{tab:natural_deployment} links smaller state variations to improved detection consistency. UGV TCS rises from 0.918 to the existing 0.949 at 1.47\% commitments; road TCS rises from 0.917 to 0.938 at 2.18\%. Selective commitment suppresses the transmission of candidate fluctuations into perception geometry, complementing cross-date pose quality.

Table~\ref{tab:jetson_execution} reports single-camera execution: 13.5 FPS, 74.1/83.9-ms mean/P95 end-to-end latency, and 8.0/9.0 ms for CAEG, with 1.47\% commitments over 2,384 valid decisions.
After warm-up, end-to-end timing on Jetson AGX Orin includes preprocessing, CAEG, SRCP, geometric-interface preparation, and frozen detection. CAEG runs every frame; commitment controls state writes. Embedded execution completes the continuous-perception chain.

\FloatBarrier

\section{Conclusion}
\label{sec:conclusion}

CalibSwitch separates observation-driven extrinsic recovery from selective commitment for continuous multimodal perception. CAEG supplies complete $\SE$ candidates; SRCP combines quality and history to adapt to persistent changes and suppress transient disturbances. Under sustained transitions, it reaches mean $E=1.46$ with 9.70\% commitments, improving frozen-detector mAP/NDS over per-frame updating by 1.43/0.92 points. Paired joint-uncertainty and matched-budget tests, two real platforms, and 13.5-FPS embedded execution establish the system value of selective geometric-state maintenance.

\bibliographystyle{IEEEtranTIE}
\bibliography{references}

\end{document}